\documentclass[runningheads]{llncs}
\usepackage{graphicx}
\usepackage{todonotes}
\usepackage{amsmath}
\usepackage{amsfonts}
\usepackage[symbol]{footmisc}
\usepackage{multirow}
\usepackage{subfigure}
\usepackage{array}
\usepackage{xspace}
\usepackage{footnote}
\usepackage{wrapfig,lipsum,booktabs}
\usepackage{upgreek}
\usepackage{tikz}
\usepackage{xcolor}
\usepackage{listings}
\usepackage{comment}
\usepackage[inline]{enumitem}

\definecolor{codegreen}{rgb}{0,0.6,0}
\definecolor{codegray}{rgb}{0.5,0.5,0.5}
\definecolor{codepurple}{rgb}{0.58,0,0.82}
\definecolor{backcolour}{rgb}{0.95,0.95,0.92}

\lstdefinestyle{mystyle}{
    commentstyle=\color{codegreen},
    keywordstyle=\color{magenta},
    numberstyle=\tiny\color{codegray},
    stringstyle=\color{codepurple},
    basicstyle=\ttfamily\small,
    breakatwhitespace=false,         
    breaklines=true,                 
    captionpos=b,                    
    keepspaces=true,                 
    showspaces=false,                
    showstringspaces=false,
    showtabs=false,                  
    tabsize=2,
    escapeinside={(*@}{@*)} 
}

\newcommand{\inlineCode}[1]{\lstinline[columns=fixed]{#1}\xspace}

\definecolor{chromeyellow}{rgb}{1.0, 0.65, 0.0}

\newcommand{\SemCloud}{\textsf{\small{SemCloud}}\xspace}
\newcommand{\onto}[1]{\textit{\textsf{\small{#1}}}}

\newcolumntype{P}[1]{>{\centering\arraybackslash}p{#1}}
\newcolumntype{M}[1]{>{\centering\arraybackslash}m{#1}}

\begin{document}

\title{Scaling Data Science Solutions with Semantics and Machine Learning: Bosch Case}

\titlerunning{Scaling Data Science Solutions with Semantics and ML: Bosch Case }

\author{
Baifan Zhou\inst{1,2,}\thanks{Baifan Zhou and Nikolay Nikolov contributed equally to this work as first authors. \email{baifanz@ifi.uio.no, nikolay.nikolov@sintef.no}} \and
Nikolay Nikolov\inst{3,1,*} \and
Zhuoxun Zheng\inst{4,1} \and
Xianghui Luo\inst{5} \and \\
Ognjen Savkovic\inst{6} \and
Dumitru Roman\inst{3,1} \and
Ahmet Soylu\inst{2} \and \\
Evgeny Kharlamov\inst{4,1}
}

\authorrunning{Zhou and Nikolov, et al.}

\institute{
Department of Informatics, University of Oslo, Norway \and
Department of Computer Science, Oslo Metropolitan University, Norway \and
SINTEF AS, Norway
\and
Bosch Center for Artificial Intelligence, Germany
\and
ACM Member, Germany \and
Department of Computer Science, Free University of Bozen-Bolzano, Italy 
}

\maketitle              % typeset the header of the contribution

\begin{abstract}
Industry 4.0 and Internet of Things (IoT) technologies unlock unprecedented amount of data from factory production, posing big data challenges in volume and variety. In that context, distributed computing solutions such as cloud systems  are leveraged to parallelise the data processing and reduce computation time.
As the cloud systems become increasingly popular, there is  increased demand that more users that were originally not cloud experts (such as data scientists, domain experts) deploy their solutions on the cloud systems. 
However, it is non-trivial to address both the high demand for cloud system users and the excessive time required to train them.
To this end, we propose SemCloud, a semantics-enhanced cloud system, that couples cloud system with semantic technologies and machine learning.
SemCloud relies on domain ontologies and mappings for data integration, and 
parallelises the semantic data integration and data analysis on distributed computing nodes.
Furthermore, SemCloud adopts adaptive Datalog rules and machine learning for automated resource configuration, allowing non-cloud experts to use the cloud system.
The system has been evaluated in industrial use case with millions of data, thousands of repeated runs, and domain users, showing promising results.\looseness=-1

\end{abstract}

\section{Introduction}
\label{sec:intro}

\begin{figure*}[t!]
% \vspace{-8ex}
	\centering
	\includegraphics[width=\textwidth]{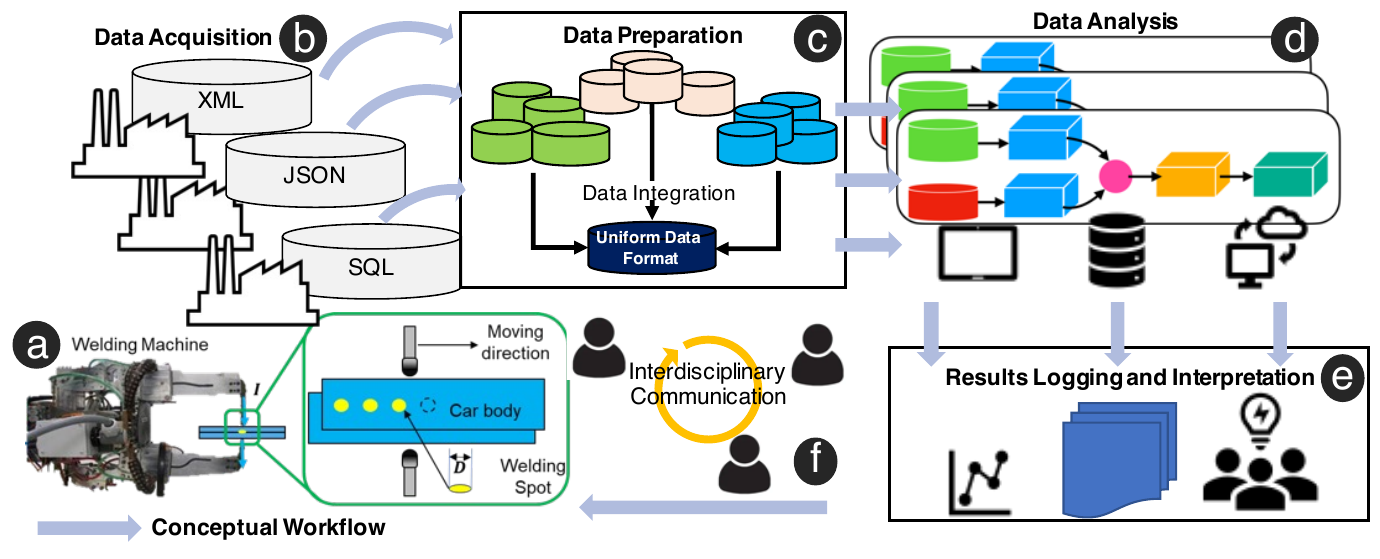}
	\vspace{-6ex}
	\caption{Data analytics development cycle exemplified on the Bosch case of welding condition monitoring. In industrial data science projects, many users are non-cloud experts (e.g., welding experts, ML experts) and want to scale their solutions on the cloud.\looseness=-1}
\label{fig:intro}
\vspace{-4ex}	
\end{figure*}

\textbf{Background.}
Industry 4.0~\cite{kagermann2015change} aims at highly automated smart factories that rely on IoT technology~\cite{ITU2012}, spanning across data acquisition, communication, information processing and actuation. 
This has unlocked unprecedented amounts of data that are generated by production systems~\cite{chand2010smart} and, thus, 
drastically increased the demand for data-driven analytical solutions and cloud technology.
We illustrate a common industrial scenario of development and deployment of data-driven solutions on cloud with a Bosch welding case\footnote{Automated welding is an impactful manufacturing process that is involved in the production of millions of cars annually, deployed world-wide at many factories. 
Data-driven analytics solutions for welding can greatly help in reducing the cost and waste in production quality. Errors in production can only be resolved by destroying newly produced cars in samples.} 
of quality monitoring in Fig.~\ref{fig:intro}:
The data from a production environment such as welding machines (a) has first to be acquired in different formats, e.g., CSV, JSON, XML (b); then they should be integrated into a uniform format (c); After that, the project team (including welding experts, data scientists, managers. etc.) wants to run data analysis on cloud infrastructures on top of the large data volumes from many factories (d);
After data analysis, these users need to discuss and log the results (e);
The whole process involves iterative and cross-domain communication between the stakeholders (f).\looseness=-1

\medskip
\noindent
\textbf{Challenges.}
From the scenario, we see that scaling data science solutions poses challenges related to dealing with the high data \textit{volume}, \textit{variety}, and more \textit{users}, namely enabling non-cloud experts to leverage cloud systems.
Indeed, industries equipped with IoT technologies produce huge volumes of production data. In the Bosch case, one factory alone produces more than 1.9 million welding records per month.
The data generated by different software versions, locations, customers have a variety of data formats, feature names, available features, etc.
Meanwhile, many users that are not cloud experts, such as domain experts and data scientists, want to deploy the data science solutions 
on the cloud.
In a standard implementation of the workflow in Fig.~\ref{fig:intro}, the project team requires extensive assistance from cloud experts, whenever they want to deploy solutions or make small changes to their solutions deployed on the cloud. 
To facilitate the adoption of cloud systems for more projects and users, one can equip all projects with some cloud experts, or launch training programs about cloud technology for all users. Both require careful planing to balance time, cost, and benefits.

\medskip
\noindent
\textbf{Our Approach.} 
To address these scalability challenges in terms of data volume, data variety, and democratising cloud systems, we propose \SemCloud: a semantics-enhanced cloud system, that scales semantic data integration and data analysis on the cloud with distributed computing, and allows non-cloud experts to deploy their  solutions. 
Our system is motivated by a use case at Bosch aiming at scaling data science solutions in welding condition monitoring (Sect.~\ref{sec:usecase}).
\SemCloud consists of semantic artefacts such as domain ontologies, mappings, adaptive Datalog rules (Sect.~\ref{sec:system}) and \textit{machine learning} (ML) that learns the parameters in the adaptive Datalog rules. \looseness=-1

In particular, 
the semantic data integration (extract-transform-load, ETL) (Sect.~\ref{sec:dataintegration}) maps diverse data sources to a unified data model and transforms them to uniform data formats.
To allow distributed ETL, \SemCloud slices the integrated data according to domain-specific data semantics (machine equipment identifiers in the Bosch case), separating the data into computationally independent subsets.  \SemCloud then  parallelises the ETL and analysis of the data slices on distributed computing nodes (Sect.~\ref{sec:distributedetl}).
Furthermore, SemCloud adopts a semantic abstraction and a graphical user interface (GUI) to democratise cloud deployment, improving transparency and usability for non-cloud experts. These include a cloud ontology that allows to encode ETL pipelines in knowledge graphs (Sect.~\ref{sec:etl-kg}), and a set of adaptive Datalog rules (Sect.~\ref{sec:mlrules}) for automatically finding optimal resource configurations. These rules are adaptive because some of their predicates are functions learnt with machine learning (ML)(Sect.~\ref{sec:ruleparalearning}).\looseness=-1

We note that the existing work on this topic addressed the cloud deployment issues only to a limited extent~\cite{youseff2008toward,ageed2020unified}, whereby they either only focus on the formal description of cloud, or on the limited adaptability of cloud systems.
\SemCloud exploits and significantly extends our previous works on ML in the context of Industry 4.0~\cite{DBLP:conf/semweb/SvetashovaZPSSM20,DBLP:journals/ws/ZhouSGSCMWK21}, and
container-based big data pipelines~\cite{nikolov_conceptualization_2021} (Fig.~\ref{fig:data_pipelines})
 by enhancing the framework with semantic artefacts and modules for specifying container-based pipelines, including pipeline step templates for containerisation and management of inter-step communication and data transmission (Sect.~\ref{sec:distributedcomputing}).

We evaluated (Sect.~\ref{sec:evaluation}) \SemCloud extensively:
the cloud deployment report to verify \SemCloud performance  on reducing computational time, with an industrial datasets of about 3.1 million welding spots;
the performance of rule parameter learning and inference based on 3562 times of repeated runs of the system.\looseness=-1

\section{Motivating Use Case: Welding Quality Monitoring}
\label{sec:usecase}
\vspace{-1ex}

In this section we discuss our motivating use case in more details, explain why scaling data science solution to large data sets  and more users is critical and discuss requirements for the  cloud system. 

\begin{figure}[t]
	% \vspace{-8ex}
	\centering
	%\hspace{-5ex}
	\includegraphics[width=\textwidth]{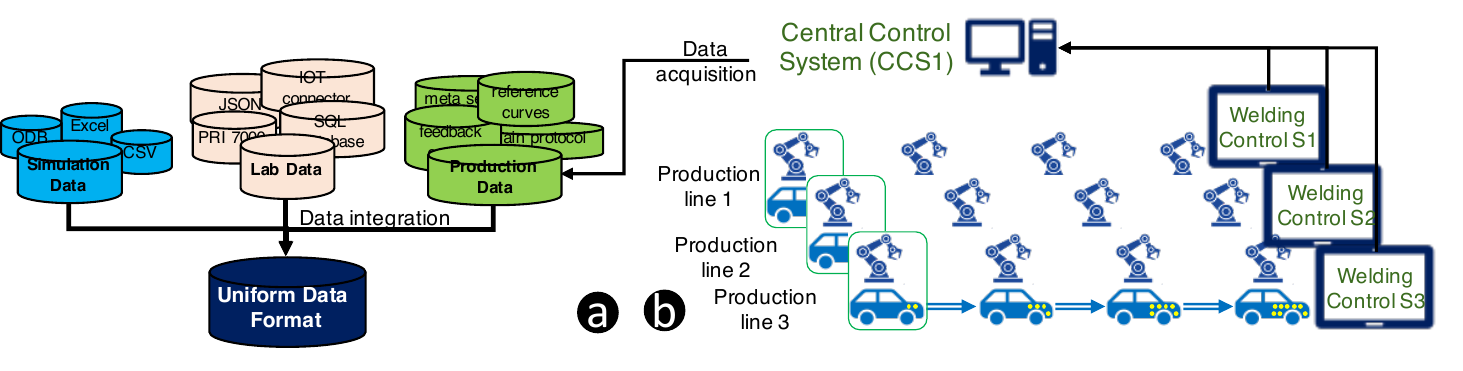}
    \vspace{-6ex}
	\caption{(a) The data variety issue. (b) The data volume issue exemplified with the production data.}
	\label{fig:semml_datavariety}
	\vspace{-4ex}
\end{figure}

\medskip
\noindent \textbf{Condition monitoring for automated welding.}
Condition monitoring refers to a group of technologies for
monitoring condition parameters in production machinery to identify potential developing faults~\cite{DIN13306_2017}.
The use case addresses one type of condition monitoring, quality monitoring (another type is machine health monitoring), for resistance spot welding at Bosch, which is a fully automated welding process that is essential for producing high-quality car bodies globally in the automotive industry. During the welding process (Figure~\ref{fig:intro}a), two welding gun electrodes press two or three worksheets (car body parts) with force,
an electric current flows through the electrodes and worksheets, generating a large amount of heat due to electric resistance. The materials in a small area between the two worksheets melt and then congeal after cooling, forming a weld nugget (or welding spot) connecting the worksheets.
The core of quality monitoring is to measure, estimate or predict some categorical or numerical quality indicators.
The diameter (Figure~\ref{fig:intro}a) of welding spots is typically used as the quality indicator of a single welding act according to industrial standards~\cite{DIN13306_2017,iso143272004}. 
Conventional practice adopts destructive methods to tear the welded car bodies apart, although they can only be applied to a small sample of car bodies, and the destroyed samples are waste and cannot be reused.
Bosch is developing data-driven solutions to  predict the welding quality, to reduce the waste and improve the coverage of quality monitoring.

 \looseness=-1

\medskip 
\noindent
\textbf{Bosch big data.}
Welding condition monitoring faces big data challenges of variety and volume. 
In terms of \textit{data variety}, 
Bosch has many data sources of different locations and conditions (Figure~\ref{fig:semml_datavariety}a). The production data alone are collected from at least four locations and three original equipment manufacturers (OEMs). These data differ in semantics and formats because of software versioning, customer customisation, as well as sensor and equipment discrepancy based on the concrete needs in the location. For example, they may be stored in various formats such as CSV, JSON, XML, etc., and may have different names for the
same variables, have some variables missing in one source but present in another, or data may be measured with different sampling rate, etc.

In terms of \textit{data volume}, 
data science models need a reasonably large amount of data to make the training meaningful and representative for the given data science tasks.
For simplicity, we consider a representative example, whereby we assume one month data are meaningful, which was confirmed by data scientists at Bosch.
In an example automobile factory responsible for manufacturing chassis (Figure~\ref{fig:semml_datavariety}b), there are 3 running production lines with a total of 45 welding machines. Each welding machine is responsible for a number of types of welding spots on the car bodies, with pre-designed welding programs. These machines perform welding operations at different speeds, ranging from one welding spot per second, to one spot per several minutes. The data related to one single welding spot consist of several protocols or databases. After integration, these data become to a set of relational tables with 263 attributes, and a simplified estimation gives that one factory produces 64.8k spots every day, and 1.944 million spots per months, which account for the production of about 432 cars. Considering an average of 125 KB data for one welding operation gives the estimation of data volume meaningful for training as 243 GB (The real amount varies and can be larger, e.g., it was estimated as 389.32 GB in one real case. Here we adopt the simplification with a similar magnitude.).
\looseness=-1

\medskip \noindent 
\textbf{Cloud deployment requirements.}
Considering  the challenges, the welding quality monitoring system should give quality estimation/prediction not with excessive response time, although the data volume is large.
In addition, the data come from various sources with diverse formats.
Moreover, industrial data science projects involve many users that are non-cloud experts (Fig.~\ref{fig:intro}). They should be equipped with tools to help deploy their data science solutions without extensive cloud expertise. The cloud infrastructure has resources of computing, memory, storage, network, etc. which need to be configured for optimised performance.
Based on the information, we derive the following requirements for the system:\looseness=-1

\begin{itemize}
[topsep=3pt,parsep=0pt,partopsep=0pt,itemsep=0pt,leftmargin=*]
    \item \textit{R1, Scalability on Data Volume:}  The system should be able to reduce the computational time significantly when processing large data volumes.
    \item \textit{R2, Scalability on Data Variety}: The system should be able to handle data variety, integrating heterogeneous data to uniform data formats.
    \item \textit{R3, Scalability on Users:} The system should improve the \textit{transparency} of the cloud system, automate resource configuration, and allow good \textit{usability} for users, especially non-cloud experts, 
\end{itemize}

\section{SemCloud: Semantics-Enhanced Cloud System}
\label{sec:system}

\begin{figure}[t!]
\vspace{-3ex}
% \hspace{-10ex}
\centering
\includegraphics[width=.8\textwidth]{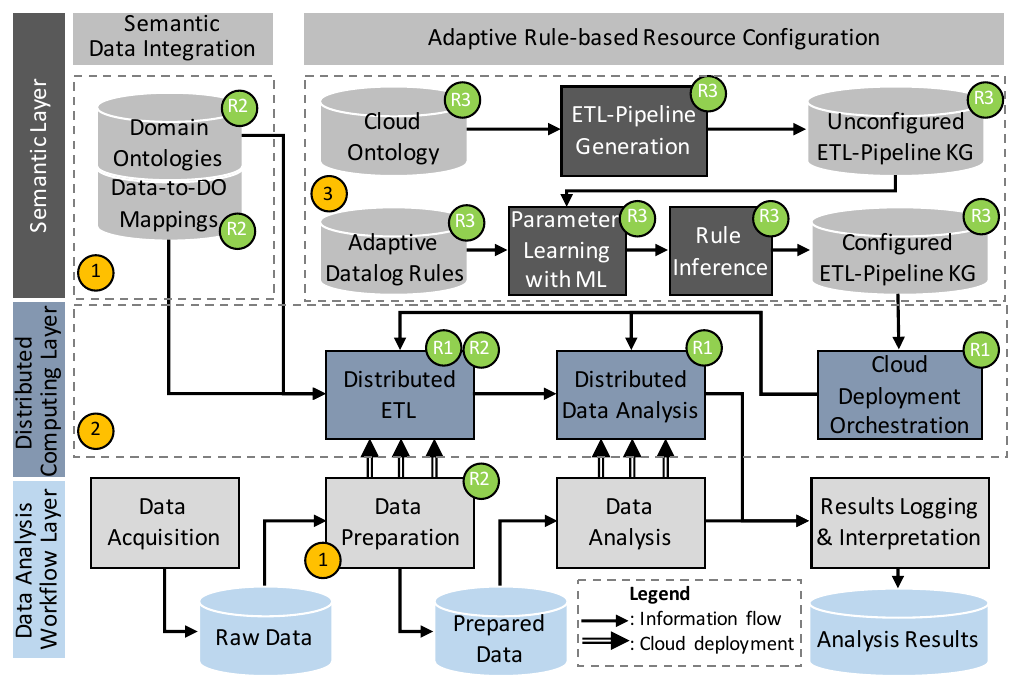}
\vspace{-2ex}
\caption{Architectural overview of \SemCloud including (1) semantic data integration; (2) distributed computing; (3) adaptive rule-based reasoning; each of which consists of a set of semantic artefacts (barrels) or modules (boxes).
\raisebox{.5pt} {\textcircled{\raisebox{-.9pt} R}} indicates the requirements the artefacts or modules intended to address.}
\label{fig:architecture}
\vspace{-2ex}
\end{figure}

To address the challenges and requirements, we propose  our \SemCloud system. We first give an architectural overview (Fig.~\ref{fig:architecture}) and then elaborate on the components.

\subsection{Architectural Overview}
The architecture of \SemCloud is shown in Fig.~\ref{fig:architecture}.
The \textit{Data Analysis Workflow Layer} adopts a common workflow: data acquisition, data preparation, data analysis, results logging and interpretation; the raw data are first acquired, then prepared for data analysis, and, finally, the analysis results are generated, including models, predictions and human interpretation.
In the data preparation stage, we employ \textit{Semantic Data Integration} (Fig.~\ref{fig:architecture}.1) that relies on domain ontologies and semantic mappings to transform diverse data sources into uniform data formats.
\SemCloud scales the data analysis workflow to the cloud by with the \textit{Distributed Computing} (Fig.~\ref{fig:architecture}.2), which includes the distributed ETL, distributed data analysis, and deployment orchestration that allocates cloud resources to the previous two modules.
\textit{Adaptive Rule-based Resource Configuration}  (Fig.~\ref{fig:architecture}.3) provides a cloud ontology and GUI for the users to encode ETL pipelines in KGs, which contain resource configuration information that is automatically reasoned by
a set of adaptive Datalog rules.
These rules consist of aggregation operations and parameterised functions, where the parameters in the functions are learned via ML.\looseness=-1

\subsection{Semantic Data Integration}
\label{sec:dataintegration}

To accommodate the diverse data sources/formats and convert all data to uniform data formats~\cite{DBLP:conf/semweb/SvetashovaZPSSM20,DBLP:journals/ws/ZhouSGSCMWK21},
we employ domain ontologies as the data models and the semantic mappings (Data-to-DO, data to domain ontology) that map diverse data sources to the data models.
In particular, the domain ontologies capture the knowledge of manufacturing processes, data, and assets. In the case of welding ontology, it is in OWL 2 QL, with 1542 axioms, which define 84 classes, 123 object properties and 246 datatype properties. The classes capture concepts such as welding operations, welding machines, welding products (spots), welding programs, sensor measurements, monitoring status, control parameters, etc.
The diverse data sources have discrepancies in data formats (e.g., CSV, JSON, XML), feature names, feature composition (some features exist in some sources but not in others), etc. All features in the different data sources of the same welding process are mapped (one-to-one mapping for each data source) to object properties (for foreign keys) or datatype properties (for attributes) in the same domain ontology. All features are renamed, and data formats are unified in one of the selected formats, usually CSV (or relational database).

\begin{figure}[t]
\vspace{-3ex}
\centering
\includegraphics[width=\textwidth]{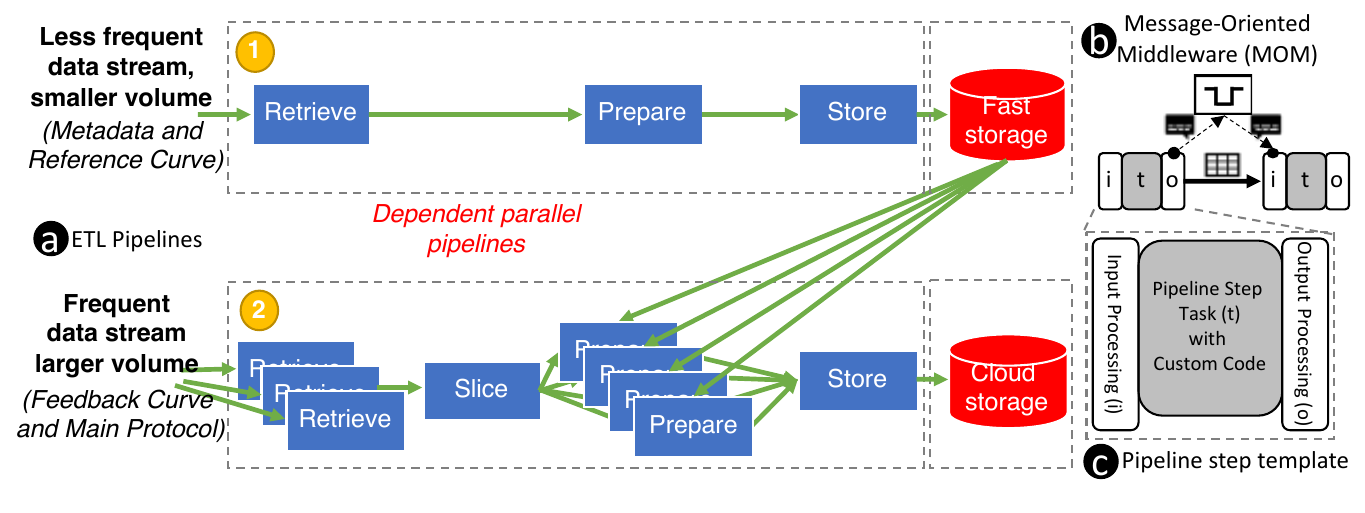}
\vspace{-7ex}
\caption{(a) Dependent parallel ETL pipelines that break down ETL into four steps: \textit{Retrieve}, \textit{Slice}, \textit{Prepare}, \textit{Store};
(b) The cloud deployment of one step in the ETL pipeline as a container, where the MOM is responsible for the communication between steps of the ETL pipeline. 
(c) Zoom in one step, we see the three parts: \textit{Input Processing}, \textit{Output Processing} and  \textit{Pipeline Step Task}.}
\label{fig:data_pipelines}
\vspace{-3ex}
\end{figure}

\subsection{Distributed ETL and Data Analysis}
\label{sec:distributedetl}

\medskip
\noindent
\textbf{Distributed ETL.}
To enable distributed ETL, we need to find a strategy that makes the ETL parallelisable, treat data streams with different updating frequencies, and handle data dependencies. 
\SemCloud achieves this by breaking down the ETL into pipelines of four steps: \textit{retrieve}, \textit{slice}, \textit{prepare}, and \textit{store} (Fig.~\ref{fig:data_pipelines}). Data retrieval constitutes the process of retrieving data from databases or online streams present in different factories and can normally be handled by a single computing node. These data are then split into subsets by the step \textit{slice} to achieve parallel processing according to data semantics that make the splitting meaningful and each subset independently processable.
In the welding use case, each subset only belongs to one welding machine,
because the data analysis of welding quality monitoring of one machine can be safely assumed to be observable or predictable with data from this particular machine, without considering other machines. In this way, the datasets are separated into subsets that are computationally independent. We then deploy the ETL stage and the subsequent two stages on the cloud system that has resources for computing, storage and networking, to reduce the overall computational time. 

An important strategy here is the hierarchically dependent parallel pipelines. We consider two types of data streams: the less frequently updated one with usually smaller volume, and the (more) frequently updated one with usually larger volume.
(1) The former one has only three ETL steps: retrieve, prepare and store, because it requires  resource (computing, storage, network) of one single cloud node and does not need slicing to parallelise. The intermediate results of this ETL pipeline are stored in a database using in-memory storage for fast query access. In the welding case, the metadata and reference curves follow this ETL pipeline.
(2) The latter one has four ETL steps because it involves the application of slicing for parallelising. The results of this ETL pipeline are stored on dedicated cloud storage. In the welding case, the processing of feedback curves and main protocol requires more resources and is implemented through this pipeline.
The ETL of these two data streams are dependent because the \textit{prepare} step of the frequent data stream must pull intermediate results of less frequent data stream.
% the metadata and reference curve pipeline. 
% his arrangement saves massive time of reading content from the cloud storage (hard disk), which is the bottleneck of the speed.

\medskip
\noindent
\textbf{Distributed data analysis.}
The key of distributed data analysis is to make assumptions of what computation is parallelisable, and split the computations into independent computing tasks.
Here the target of data analysis is to predict the welding quality quantified by quality indicators such as spot diameters or Q-Values~\cite{DBLP:conf/cikm/ZhouSBPMK20,DBLP:journals/ws/ZhouSGSCMWK21}.
The tasks include both classification (good or bad quality), regression (diameter values~\cite{zhou2018comparison} or Q-Values), and forecasting~\cite{zhou2022machine} (predicting quality in the future).  In practice, the latter two are preferred by domain experts because they provide more insights than a simple  classification.

Both classic methods (feature engineering with e.g., linear regression) and deep learning (LSTM networks) are employed.
We developed and tested various ML models~\cite{DBLP:conf/cikm/ZhouSBPMK20,zhou2022machine}. 
We used model performance for tuning the hyper-parameters and considered both model performance and adoption difficulty for selecting the best models~\cite{DBLP:journals/ws/ZhouSGSCMWK21}.
These models  take input features such as sensor measurements, monitoring status and control parameters and predict the quality indicators.
The training was done with various regimes~\cite{DBLP:phd/dnb/Zhou21}: the ground truth training data included simulation data, lab data, historical production data; the validation data were subsets of the training data for selecting hyper-parameters; test data were both of the same welding machines or different machines (testing transferability).
According to domain knowledge, we assume that the interplay between welding machines to be only marginally significant and that it is safe to predict the welding quality of one welding machine only by using information of this welding machine. This assumption has been verified and obtained a prediction error of about 2\%~\cite{DBLP:conf/cikm/ZhouSBPMK20}.
Thus, the data analysis on data of each welding machine can be performed independently if each subset contains all data of one machine. %This is achieved already in data slicing as explained in the paragraph of \textit{Distributed ETL}. 
\looseness=-1

% Thus, each cloud node runs independent data analysis models and parallelise the data analysis.
% The interplay between the welding machines could be that the machines  operate on the same set of car bodies and in the same environment. 
% We consider the interplay between welding machines to be marginally significant 
% because welding quality should be insensitive to car bodies and environment according to domain knowledge.
% For more information we refer the readers to our previous works~\cite{DBLP:conf/cikm/ZhouSBPMK20,DBLP:journals/ws/ZhouSGSCMWK21}, which elaborated and extensively verified this part.

\medskip
\noindent
\textbf{Cloud Deployment Orchestration.}
\label{sec:distributedcomputing}
% The voluminous heterogeneous data need to be integrated by ETL pipelines.
% To specify, dynamically configure and manage the ETL pipelines, we adopt a state-of-the-art approach for Big Data pipelines~\cite{nikolov_conceptualization_2021}. 
To orchestrate the distributed computation,
\SemCloud
encapsulates  ETL  steps or data analysis
as containers and runs the containers independently and in parallel,  allowing for deploying multiple instances of the same ETL step or data analysis \cite{nikolov_conceptualization_2021}. Each instance is implemented by a template composed of three main parts: \textit{Input Processing}, \textit{Pipeline Step Task}, and \textit{Output Processing} (Figure~\ref{fig:data_pipelines}c).
The \textit{Input Processing} fetches data from remote sources and moves the data to the step workspace.
The \textit{Pipeline Step Task} wraps custom code to process the fetched data.
The \textit{Output Processing} component delivers the processed data to a specific destination, notifies that they are available for the next steps, and clears up  temporary and input data from the step workspace. 
Configuration and attributes of a pipeline step can be expressed as parameters and injected at deployment time. 
The communication between the steps is handled by Message-oriented Middleware (MOM)~\cite{albano2015message} (Figure~\ref{fig:data_pipelines}b), so that the consecutive steps do not need to run simultaneously for interaction, ensuring temporal decoupling. None of the sequential steps needs to know about the existence of other steps or their scaling configuration, thus achieving space decoupling.
Therefore, it is possible to assign more instances to bottlenecked pipeline steps that are more computationally heavy and reduce the overall processing time.

% \subsection{Adaptive Rule-based Reasoning for Resource Allocation}
\subsection{ETL Pipeline Generation}
\label{sec:etl-kg}

\begin{figure}[t]
% \vspace{-3ex}
\centering
\includegraphics[width=\textwidth]{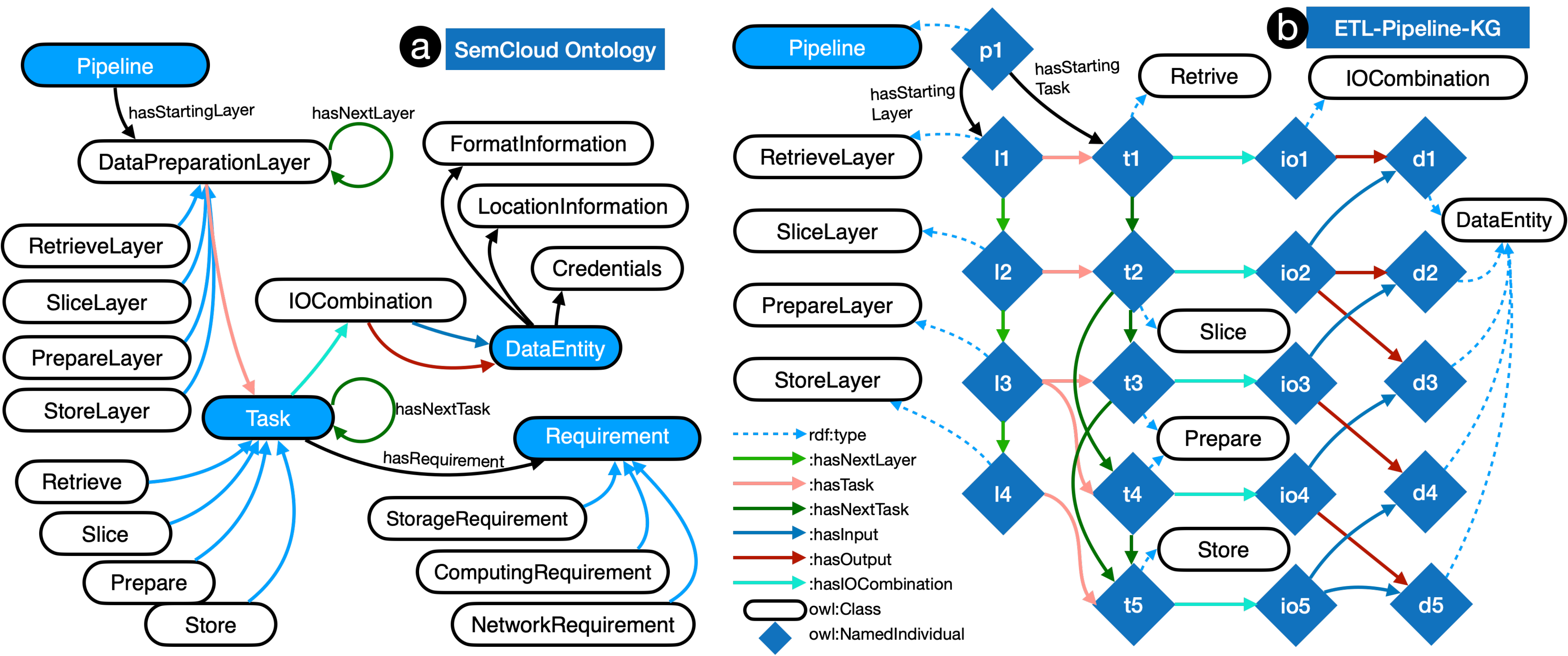}
\vspace{-5ex}
\caption{(a) Schematic illustration of the SemCloud ontology. (b) Partial illustration of a KG for \texttt{ETL-Pipeline}.}
\label{fig:ontology_etl_pipeline}
\vspace{-3ex}
\end{figure}

% \medskip 
\noindent \textbf{Cloud Ontology.}
\SemCloud provides the users GUI to construct ETL pipelines and encode them into knowledge graphs, based on a SemCloud ontology (Figure~\ref{fig:ontology_etl_pipeline}a).
The ontology \onto{SemCloud} is written in OWL 2, and consists of 20 classes and 165 axioms. It has three main classes: \onto{DataEntity}, \onto{Task}, \onto{Requirement}. \onto{DataEntity}  refers to any dataset  to be processed;
\onto{Task} has sub-classes that represents the four types of tasks in the data preparation: \textit{retrieve}, \textit{slice}, \textit{prepare}, and \textit{store};
and \onto{Requirement} that describes the requirements for computing, storage and networking resources.

\medskip \noindent
\textbf{ETL Pipeline Generation in KGs.}
We now illustrate the generation of ETL pipelines in knowledge graphs with the example in Figure~\ref{fig:ontology_etl_pipeline}b.
The data for welding condition monitoring have multiple levels of updating frequencies, which should be accommodated by the ETL pipelines.
For example,
data that are generated for each welding operation are updated after each welding operation, and thus are updated very frequently (about one second for one operation). 
For these data, the users construct an ETL pipeline \texttt{p1} with four layers (via GUI).
Firstly, data are ``retrieved'' from the welding factories. Thus, the layer \texttt{l1} is of type \onto{RetrieveLayer}, and has the task \texttt{t1} of type \onto{Retrieve}. The task \texttt{t1} has an IO handler \texttt{io}, which has an output \texttt{d1} of type \onto{DataEntity}.
Then the data are read in by a task \texttt{t2} of type \onto{Slice}, 
and ``sliced'' into smaller pieces \texttt{d2}, \texttt{d3}.
These slices are input to different computing nodes to do tasks \texttt{t3} and \texttt{t4}  of type \onto{Prepare}. Finally, all prepared data entities are stored by \texttt{t5} of type \onto{Store}.

\subsection{Adaptive Datalog Rules Inference for Resource Configuration}
\label{sec:mlrules}

% \begin{figure}[t]
%      \centering
%      \includegraphics[width=\textwidth]{Figures/rules.pdf}
%      \caption{Caption}
%      \label{fig:my_label}
% \end{figure}

% The traditional way : 
Obtaining an optimised cloud configuration is not a trivial task. Cloud experts typically try different configurations by testing the system  with various settings and use the system performance under these test settings as heuristics to manually decide the cloud configurations.
In \SemCloud, we design a set of declarative adaptive rules written using logical programming to make the cloud configurations explicit, automated, less error-prone where the system optimisation is done with help of external functions learned with ML, such that the rules can be also used  by non-cloud experts.

To this end, \SemCloud uses adaptive rules in
Datalog with aggregation and calls to external predicates learned by ML (they are adaptive because the function parameters are learned, see Sect.~\ref{sec:ruleparalearning}). 
In particular, 
we consider non-recursive rules of the form 
$A \leftarrow B_1, \ldots, B_n$
where $A$ is a head of rule (the consequence of the rule application) 
and $B_1, \ldots, B_n$ are either 
 predicates that apply join or aggregate function 
that filters out the results.
For the theouploary of Datalog we refer 
to~\cite{leone2002dlv,DBLP:journals/corr/cs-AI-0404011,abiteboul1995foundations,paramonov2013asp}. 
In the following we provide some example rules and explain their logic.

We have six different Datalog programs 
(set of rules) that run independently
and that are divided into 
three steps:
\begin{enumerate*}[label=(\roman*)]
   \item graph extraction rules that populate rule predicates by extracting information from the ETL-pipeline KGs (e.g., $\textit{rule}_0$)
   \item resource estimation rules that estimate the resource consumption for the given pipeline if there is only one computing node (assuming infinitely large nodes, e.g., $\textit{rule}_2$) 
   \item resource configuration rules that find the optimal resource allocation in distributed computing the given pipeline (e.g., $\textit{rule}_3$).
\end{enumerate*}
% such that first apply graph extraction rules, then estimation rule and finally resource allocation rules.

\smallskip
\noindent
\textbf{Graph extraction rules.}
These rules populate the predicates so that these predicates will be used for the resource estimation and configuration.
The $\textit{rule}_0$   exemplifies populating the predicate \inlineCode{subgraph1} that is related to the ETL pipeline \inlineCode{p}.

\begin{lstlisting}[language=Prolog]
subgraph1(p,n,v,ms,mp,ssl,spr,sst) (*@ $\leftarrow$ @*) ETLPipeline(p),
   hasInputData(p,d), hasVolume(d,v), hasNoRecords(d,n)
   hasEstSliceMemory(p,ms), hasEstPrepareMemory(p,mp)
   hasEstSliceStorage(p,ssl), hasEstPrepareStorage(p,spr)
   hasEstStoreStorage(p,sst) (*@ \hfill (${rule}_0$) @*)
\end{lstlisting}
Similarly, we have rule $\textit{rule}_1$ that creates \inlineCode{subgraph2(p,n,v,ms,mp,ts,tp,nc,ns,mrs,mrp,mode)}.

% Graph extraction rules, such as $\textit{rule}_0$ below, 
% populate predicates using the information from the underlying KG.

% \begin{align*}
% & \textit{rule}_0: \comp{subgraph1(p,n,v,ms,mp,ssl,spr,sst)}
% \leftarrow    \\
%   & \comp{ETLPipeline(p), hasInputData(p,d), hasVolume(d,v),} \\
%   & \comp{hasNoRecords(d,n),
% hasEstSliceMemory(d,ms),}
% \\
%   & \comp{hasEstPrepareMemory(d,mp),
% hasVolume(d,v), hasNoRecords(d,n)},
% \\
%   & \comp{hasEstSliceMemory(d,ms),
% hasEstPrepareMemory(d, mp)},
% \\
%   & \comp{hasEstSliceStorage(d,ssl),
% hasEstPrepareStorage(d,spr)},
% \\
%   & \comp{hasEstStoreStorage(d,sst)}.
% \end{align*}

% \resizebox{.9\linewidth}{!}{
%    \begin{minipage}{\linewidth}
% \begin{lstlisting}[language=Prolog]
% subgraph1(p,n,v,ms,mp,ssl,spr,sst) (*@ $\leftarrow$ @*)
%    ETLPipeline(p), hasInputData(p,d), 
%    hasVolume(d,v), hasNoRecords(d,n)
%    hasEstSliceMemory(d,ms), hasEstPrepareMemory(d,mp)
%    hasEstSliceStorage(d,ssl), hasEstPrepareStorage(d,spr)
%    hasEstStoreStorage(d,sst)
% \end{lstlisting}
%    \end{minipage}
% }

% \inlineCode{subgraph2(p,n,v,ms,mp,ts,tp,nc,ns,mrs,mrp,mode)}
% populates predicate that we use in the next steps.

\smallskip
\noindent
\textbf{Resource estimation rules.}
These rules are used to estimate required resources assuming one computing node. For example, $\textit{rule}_2$ estimates the required slice memory (\inlineCode{ms}), prepare memory (\inlineCode{mp}), slice storage (\inlineCode{ssl}), prepare storage (\inlineCode{spr}), and the store storage (\inlineCode{sst}). The rule then stores these estimations in the predicate
\inlineCode{estimated_resource}.

\begin{lstlisting}[language=Prolog]
estimated_resource(p,ms,mp,ssl,spr,sst) (*@$\leftarrow$@*)
    subgraph1(p,n,v,ms,mp,ssl,spr,sst),
    ms=@func_ms(n,v), mp=#avg{@func_mp(n,v,ms,i):range(i)},
    spr=#avg{@func_spr(n,v,ssl,i):range(i)}, 
    ssl=@func_ssl(n,v), sst=@func_sst(n,v,ssl,spr) (*@ \hfill (${rule}_2$) @*)
\end{lstlisting}

% \begin{verbatim}
% rule_3:
% estimated_resource_conf(p,ms,mp,ssl,spr,sst) <-
% subgraph1(p, n, v, ms, mp, ssl, spr, sst)
% ms = @func_ms(n, v), mp = #avg { @func_mp(n, v, ms,i): range(i) }, 
% ssl = @func_ssl(n, v), spr = #avg { @func_spr(n, v, ssl) : range(i) },
% sst = @func_sst(n, v, ssl, spr)
% \end{verbatim}

\noindent where  \inlineCode{@func_ms}, \inlineCode{@func_ssl}, 
\inlineCode{@func_sst}, etc. are parameterised ML functions whose parameters are
learnt in the \textit{rule parameter learning} (Sect.~\ref{sec:ruleparalearning}). In the implementation, 
those are defined as external built-in functions 
that are called in the grounding phase of the program
and then are replaced by concrete values~\cite{leone2002dlv,DBLP:journals/corr/cs-AI-0404011}.
We also have other resource estimation rules that  estimate  other resources, such as  
CPU consumption. 

\smallskip
\noindent
\textbf{Resource configuration rules.}
These rules find the optimal cloud configurations based on the estimated cloud resource.
${rule}_3$ is an example for
% Those rules have are organised in two steps. First 
% we join data obtained in the previous step into 
% an auxiliary predicate 
% \inlineCode{configured_resource_aux}
% as in ${rule}_3$, then the optimal resource configurations are inferenced as in ${rule}_4$. 
% These are examples for 
deciding the slicing strategy and the storage strategy, and finding the optimal resource configuration such as the chuck size 
(\inlineCode{nc}), slice size (\inlineCode{ns}), memory reservation for \textit{slice} (\inlineCode{mrs}) and for \textit{prepare} (\inlineCode{mrp}). In essence, $rule_3$ stipulates that if the maximum of estimated slice memory (\inlineCode{ms}) and prepare memory (\inlineCode{mp}) is greater than a given threshold (\inlineCode{c1*nm}), and the maximum of estimated slice storage (\inlineCode{ssl}), prepare storage (\inlineCode{spr}), and store storage (\inlineCode{sst}) is smaller than (or equal to) another threshold (\inlineCode{c2*ns}), then the chosen strategy for the given pipeline is \textit{slicing} (thus \inlineCode{nc} and \inlineCode{ns} are computed), and \textit{fast storage} (\inlineCode{fs}, where the thresholds are calculated from cloud attributes.

\begin{lstlisting}[language=Prolog]
configured_resource(p,nc,ns,fs,mrs,mrp) (*@$\leftarrow$@*)
    subgraph2(p,n,v,ms,mp,ts,tp,nc,ns,mrs,mrp,mode),
    estimated_resource(p,ms,mp,ssl,spr,sst),
    CloudAttributes(c,c1,c2,c3,nm,ns,fs,cs),
    #max{ms,mp} > (c1 * nm), #max{ssl,spr,sst} <= (c2 * ns),
    nc = @func_fs_1(n,v,ts,tp), ns = @func_fs_2(n,v,ts,tp),
    mrs = #min{ms, #max{@func_ss(n,v,nc,ns), c3*ms}},
    mrp = #min{mp, #max{@func_pn(n,v,nc,ns), c3*mp}} (*@ \hfill (${rule}_3$) @*)
\end{lstlisting}

\subsection{Rule Parameter Learning with Machine Learning}
\label{sec:ruleparalearning}
% Step 1: learn rule parameters from pilot running statistics.

% Step 2: inference the resource allocation.

The functions in the adaptive rules are in the form of ML models. The \textit{resource estimation rules} are  selected from the best model resulting from training three ML methods and the pilot running statistics. These three ML methods are \textit{Polynomial Regression (PolyR)} (Eq.~\ref{eq:polyr}),
\textit{Multilayer Percetron (MLP)} (Eq.~\ref{eq:mlp}),
and \textit{K-Nearest Neighbours (KNN).} (Eq.~\ref{eq:knn}). We selected these three methods because they are representative classic ML methods suitable for the scale of the pilot running statistics.
PolyR transfers the input features ($x_i,i\in\{1,2,...,n\}$, $n$ is the number of input features) to a series of polynomial vectors ($[1, x_i,  x_i^2, ... x_i^m]$, $m$ is the highest degree), and then constructs a predictor by multiplying a weight matrix ($\mathbf{W}\in \mathbb{R}^{m \times n}$).
MLP consists of multiple layers of perceptrons, 
where each perceptron applies the \textit{ReLu} function to the weighted sum of all neuron outputs of the previous layer plus the bias terms $\mathbf{W}^{(l-1)}\mathbf{h}^{(l-1)} + \mathbf{b}^{(l-1)}$.
For a given data $i$ whose output feature $y_i$ is to be predicted, KNN  finds its k samples (s, consisting of pairs of input $\mathbf{x}_s$ and output  $y_s$) that are most similar to $i$ (the k nearest neighbours $\mathcal{N}_k$) in the training data,  and uses a weighted sum (the reciprocal of distance $d(s,i)$) of the output features $y_s$ in $\mathcal{N}_k$ as the estimation.\looseness=-1
% of the given instance $i$, where the similarity between $s$ and $i$ is defined by a distance  function $d(s,i)$ (e.g., the Euclidean distance).\looseness=-1

The \textit{resource configuration rules} are trained with the same three ML methods and with optimisation techniques such as Bayesian optimisation or grid search.
For example, the functions \inlineCode{@func_fs_1} and \inlineCode{@func_fs_2} that find the optimal chuck size (\inlineCode{nc}) and slice size (\inlineCode{ns}) are trained by finding the arguments of (\inlineCode{nc,ns}) for  the minimal total computing time ($t_{\texttt{total}}$)

\[
\vspace{-2ex}
\texttt{nc},\texttt{ns} = \underset{\texttt{nc},\texttt{ns}}{\arg\min} \; {t_{\texttt{total}}} =  \underset{\texttt{nc},\texttt{ns}}{\arg\min} \: f(\texttt{v}, \texttt{n},\texttt{nc},\texttt{ns},t_{\texttt{slice}},t_{\texttt{prepare}})
% \vspace{-2ex}
\]

% \begin{lstlisting}[language=Prolog]
% nc, ns = (*@$\underset{\texttt{nc},\texttt{ns}}{\arg\min} \; {t_{\texttt{total}}} = \underset{\texttt{nc},\texttt{ns}}{\arg\min} \: f$@*) (v,n,nc,ns,(*@$t_{\texttt{slice}},t_{\texttt{prepare}}$ @*))
% \end{lstlisting}

\begin{table}[h!]
\vspace{-8ex}
\resizebox{\textwidth}{!}{
\setlength{\tabcolsep}{.04\textwidth}
\begin{tabular}{p{.2\textwidth}p{.4\textwidth}p{.25\textwidth}}
\begin{equation}\label{eq:polyr}
\begin{aligned}
    % & \mathbf{y}^{(0)} = \mathbf{x}\\
    \mathbf{x}_i &= [1, x_i,  x_i^2, ... x_i^m]^{\mathbf{T}}\\
    \hat{y}_i &=\sum { }_i{ \mathbf {W} \mathbf{x}_i }\\
    err &= || \hat{\mathbf{y}} - \mathbf{y} ||^2
    % \mathbf{y} = \sum\sum_{n=1}^{\infty} 2^{-n} \mathbf{W} \mathbf{x} + \mathbf{b}
\end{aligned}
\end{equation}
&
\begin{equation}\label{eq:mlp}
\begin{aligned}
    \mathbf{h}^{(0)} &= \mathbf{x} = [x_1,x_2,...,x_n]^{\mathbf{T}}\\
    \mathbf{h}^{(l)} &= ReLu(\mathbf{W}^{(l-1)}\mathbf{h}^{(l-1)} + \mathbf{b}^{(l-1)})\\
    \hat{y} &= ReLu(\mathbf{W}^{(L-1)}\mathbf{h}^{(L-1)} + {b}^{(L-1)})\\
    % & 1/m\sqrt{ \sum_{i=1}^m (y_i- \hat{y}_i)^2}
\end{aligned}
\end{equation}
&
\begin{equation}\label{eq:knn}
\begin{aligned}
    s &= (\mathbf{x}_s,y_s)\\
    \mathcal{N}_k &= \{ s | d(s,i) \le d_k \}\\
    d(s,i) &= ||\mathbf{x}_s - \mathbf{x}_i||\\
    \hat{y}_i &= 
    % \sum_{s \in \mathcal{N}_k}{w_sy_s}
    \mathbf{w}\mathbf{y}_s, s \in \mathcal{N}_k
    % \frac{1}{m}\sum_{s \in \mathcal{N}}{wy_s}
\end{aligned}
\end{equation}
\end{tabular}
}
\vspace{-10ex}
\end{table}

\section{Implementation and Evaluation}
\label{sec:evaluation}

We implemented our system with a front-end GUI based on Angular, HTML/CSS, and a back-end system based on ASP.NET Core, JavaScript, Python and DLV system~\cite{dlvhex,eiter2018dlvhex}.
The GUI adopts the common design pattern of Model-View-Controller and has a RESTful API that handles the requests and responses between the front-end and back-end.\looseness=-1

The evaluation consists of 
(\ref{sec:cloudeva}) cloud deployment report, verifying to what extent \SemCloud reduces computational time for semantic ETL (R1, R2);
and  (\ref{sec:mleva}) rule parameter learning and inference, validating whether the rule parameter learning and inference is scalable (R1) and accurate, so that the non-cloud experts can use \SemCloud with confidence (R3).

\subsection{Cloud Deployment Report}
\label{sec:cloudeva}

%\medskip 
\noindent \textbf{Data Description.}
To determine whether \SemCloud reduces computational time, we use a dataset of 3 production lines for one month. The dataset is
anonymised and simulated based on a welding factory in Germany. 
We simulated the dataset because it allows the freedom of evaluating settings and the information of real data is subject to a non-disclosure agreement.
% To test the limit of our SemCloud system, we select a dataset consisting of anonymised and synthesised data. 
One production line has 10 - 20  machines, amounting to 45 machines in total. Each  machine performs welding operations at a different speed, ranging from 1 spot/second, to 1 spot per several minutes (due to maintenance time, delay time, and various situations). The total amount of data are 389.42 GB, which represent 3.1 million spots, estimated to be related to 692.3 cars

\medskip \noindent \textbf{Deployment Setting.}
We deployed the \SemCloud system on an infrastructure of 7 computing instances connected by a network that were managed by a Rancher container orchestrator~\cite{rancher}. 
% These instances have shared cloud storage, 
We adopt the automatic setting, whereby resource configurations are provided via adaptive Datalog rules and the Rancher system automatically assigns containers to resources according to the configuration.

\begin{figure}[t]
% \vspace{-8ex}
	\centering
	\small	
	\includegraphics[width=\textwidth]{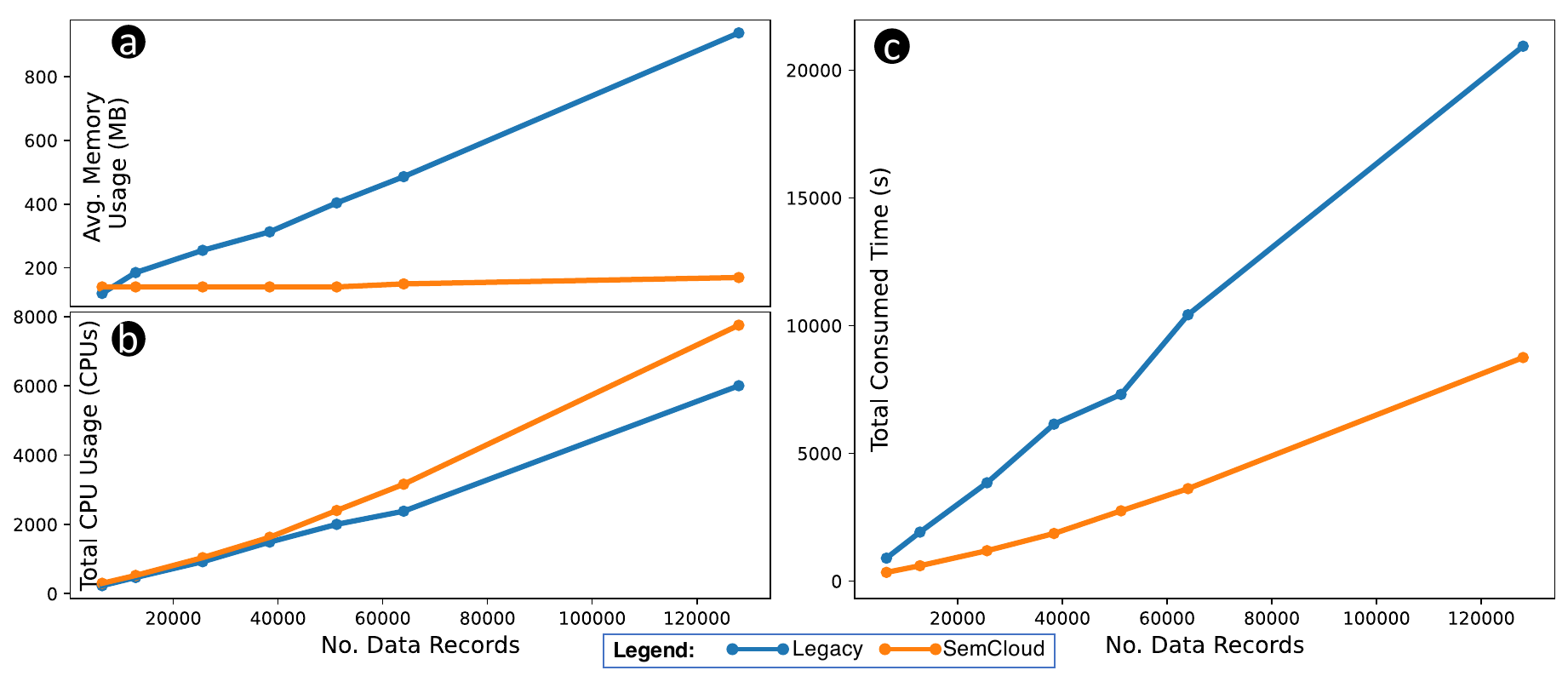}
	\vspace{-5ex}
	\caption{(Performance comparison by (a) memory usage, (b) total CPU usage (integrated over time), (c) consumed time: \SemCloud significantly reduces the memory usage and consumed time (by about 50\%), and uses slightly more total CPU, compared to the \textit{legacy} solution (Without SemCloud). X-axis: processed input data volume.}
\label{fig:deploymentresults}
\vspace{-2ex}	
\end{figure}

\medskip \noindent \textbf{Performance Comparison.}
We demonstrate the performance comparison between the ETL processing with the legacy system (without \SemCloud) and with our \SemCloud system (Figure~\ref{fig:deploymentresults}). 
The legacy system is comprised of a integrated software that performs both the preparation of the metadata/reference curves and the processing of the feedback curve and main protocol data. 
The legacy system was deployed in a node that meets the total requirements for the experimental data to monitor the resource usage.
% (the same node is used non-exclusively in the distributed experiment as one of the potential targets for deployment). 

It can be seen that the the memory usage of the computing instance for the legacy solution increases monotonously along the processed input data volume, while \SemCloud requires almost zero increase of memory allocation, which means \SemCloud can deploy the ETL process on many computing instances with no extra memory demand.
Figure~\ref{fig:deploymentresults}b shows \SemCloud requires slightly more CPU power. This is expected and understandable, because the distributed deployment consumes more computing power per unit of time, but decreases the overall computing time. The latter is confirmed by Figure~\ref{fig:deploymentresults}c: as the input data volume increases the reduced computing time brought by \SemCloud becomes increasingly significant (R1).

\subsection{Evaluation of Rule Parameter Learning and Inference}
\label{sec:mleva}

\noindent \textbf{Pilot Running Statistics}.
To verify the scalability and accuracy of the rule parameter learning and inference, we gather pilot running statistics, train and test the ML functions in \SemCloud.
We run \SemCloud repeatedly 3562 times with different
 sizes of subsets of the welding dataset in Sect.~\ref{sec:cloudeva} and gather pilot running statistics. These statistics include data information, e.g., input data size, different configurations, e.g., slice size, and recorded resource consumption e.g., memory consumption, mCPU consumption.

\medskip \noindent \textbf{Experiment Setting.}
We split the pilot running statistics so that 80\% are for rule parameter training and 20\% for rule testing and inference.
% as training set, for learning the rules, and 20\% as the test set ($\mathcal{D}_{test}$) for verifying the learned rules. 
Three ML models are trained and tested: \textit{PolyR}, \textit{MLP}, and \textit{KNN}. We adopt a grid search strategy for hyper-parameter tuning. The final hyper-parameter are, PolyR: 4 degree; MLP:  2 hidden layers with neurons 10 and 9;  KNN: 2 neighbours.

\medskip \noindent \textbf{Performance Metrics.}
We use several performance metrics:
\textit{normalised mean absolute error} (\textit{nmae})~\cite{chai2014root} to measure prediction accuracy, minimal training data amount (Min. $|\mathcal{D}_{train}|$ for yielding satisfactory results, optimisation time (Opt. time), learning time and inference time.
Intuitively, \textit{nmae} reflects the scale-independent average prediction error. It is computed as \textit{mean absolute error} normalised by the mean value of the configuration $\bar{c}$: 
$\textrm{\textit{nmae}} = mae/\bar{c}$.
\textit{nmae} reflects the mean absolute error between the ground truth configuration $c$ and the predicted configuration $\hat{c}$: 
$mae = \frac{1}{|\mathcal{D}_{test}|} \sum_{ \mathcal{D}_{test}} |c - \hat{c}|$, 
and $\bar{c} = \sum_{ \mathcal{D}_{test}} c/|\mathcal{D}_{test}|$. We normalise $mae$ because
its scale is dependent on the variable for which we calculate \textit{mae}. If it is divided by the mean value of the variable, it becomes \textit{nmae} which is scale-independent.\looseness=-1

\begin{figure}[t]
% \vspace{-8ex}
	\centering
	\small	
	\includegraphics[width=1\textwidth]{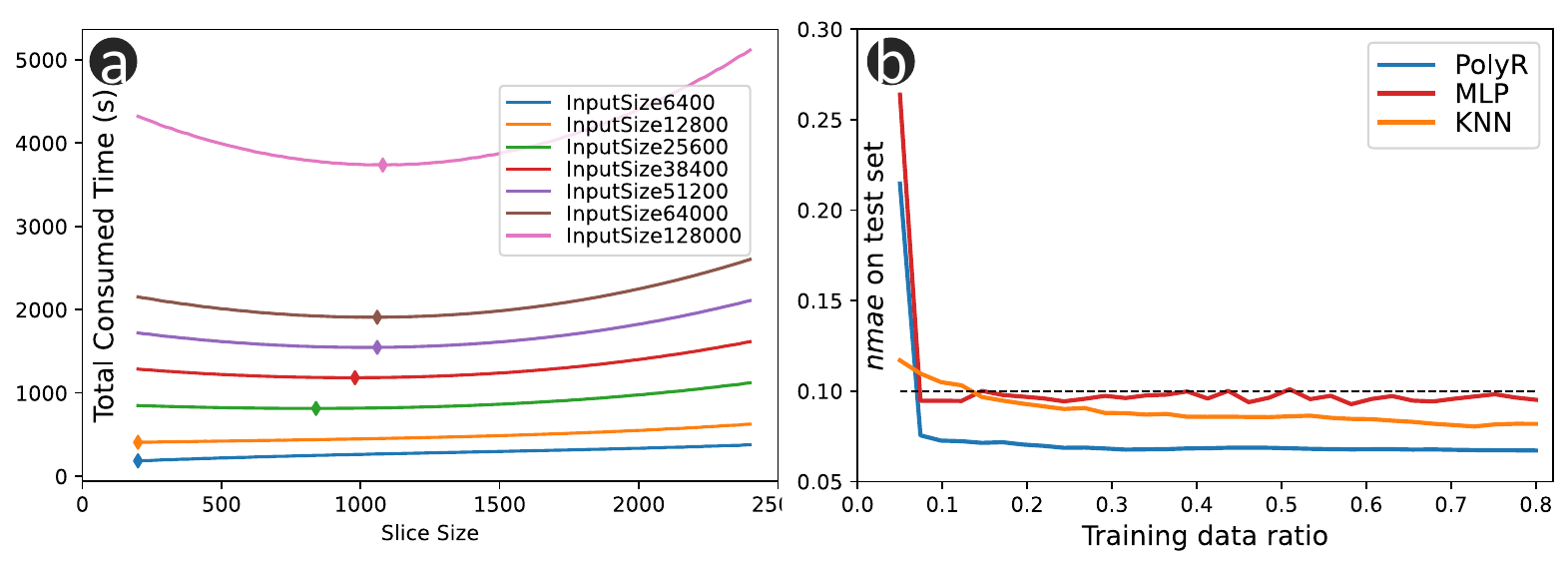}
	\vspace{-6ex}
	\caption{(a) Optimisation to find the best slice size for the least time, when chunk size is fixed. (b) Comparing ML methods to find the minimal training data}
\label{fig:optimisation_trainratio}
\vspace{-4ex}	
\end{figure}

\begin{wraptable}{r}{.5\textwidth}
\vspace{-8ex}
\caption{Parameter learning and reasoning results, recorded on Intel Core i7-10710U.}
\label{tab:ruleresults}
\setlength{\tabcolsep}{1mm}
\resizebox{.5\textwidth}{!}{
\begin{tabular}{lccc}
\toprule  
Metric & PolyR & MLP & KNN \\ \hline
\textit{nmae} & 0.0671 & 0.0947 & 0.0818 \\
Min. $|\mathcal{D}_{train}|$ & 7.42\% & 50.97\%  & 10.00\% \\
Opt. time & 1.12s & 174.32s & 7.25s \\
Learning time & 20.82ms & 120.31ms & 27.52ms \\
Inference time & $<$1.00ms & $<$1.00ms & $<$5.00ms \\ \hline
\end{tabular}}
\vspace{-3ex}
\end{wraptable}
\medskip \noindent \textbf{Learning and Inference Results.} The optimisation results of \textit{slice size} shows we can configure it to find a ``sweet spot'' to minimise the total consumed time (Figure~\ref{fig:optimisation_trainratio}a).
The performance of rule learning and inference is 
shown in Table~\ref{tab:ruleresults} (R3). 
The learning time is the time it takes to train the models.  
The inference time includes the model inference time and the rule reasoning time.
It can be seen that the PolyR has the best prediction accuracy, requires the least training data, and consumes the least time. Therefore, PolyR generates the best results and is selected for the use case. We presume the reason is that PolyR works better with small amounts of and not very complex data (3562 repeated running statistics). We can see MLP is not very stable (Figure~\ref{fig:optimisation_trainratio}b). This is  due to the random initialisation effect of MLP.

\section{Discussion on General Impact and Related Works}
\label{sec:relatedwork}
\vspace{-1ex}

\noindent 
\textbf{Uptake for the cloud community.}
\SemCloud is an attempt for democratising cloud systems for non-cloud experts. We hope to inspire research and a broad range of users who are pursuing scaling data science solutions on the cloud, but are impeded by the long training time for acquiring cloud expertise.
Providing a dynamically scalable (on a step and pipeline level), general-purpose solution for big data pipelines on heterogeneous resources  that a broad audience can use is an open research topic~\cite{barika_orchestrating_2019,buyya_manifesto_2018}. 
The currently available multitude of tools for big data pipeline management only partially addresses these issues as shown in \cite{matskin_survey_2021}. Data pipeline management approaches in literature such as \cite{gerlach_skyport-container-based_2014,qasha_dynamic_2016,alaasam_scientific_2018} also partially address the issue but either specifically support a knowledge domain (scientific workflows, ad-hoc languages), fail to address issues related to individual step scalability, or are not well-suited for dynamic long-running jobs. 
Other works that touch that topic~\cite{tan2020lstrap,zhao2021ufc2,kumar2019demand,mulfari2015computer} also do not address the automatic resource allocation issue and are not designed for non-cloud experts.
Our  approach tackles these issues and the presented principles should be easy to reproduce.

\smallskip
\noindent 
\textbf{Uptake in terms of semantic technology.}
We open-source our cloud ontology~\cite{semcloudontology}.
% \footnote{\url{https://github.com/nsai-uio/SemCloud}}.
We hope this ontology can facilitate research of semantic technology in the scalability challenges, that the tenets of explicit, transparent, and shared knowledge can advance in  the practice in academia and industry.
We developed it as we did not find a suitable ontology for our challenges.
Past works about cloud ontologies focus more on describing the different layers and components~\cite{youseff2008toward,ageed2020unified}, services~\cite{tahamtan2012cloud}, functional or non-functional features and the interaction between the layers~\cite{al2020cloudfnf}. They cover the cloud tasks and resource allocation, but to a limited extent. There exist other works about the resource management topic~\cite{castane2018ontology,ma2011ontology,zhang2016particle}, but they do not provide mechanism or 
reasoning for adaptive and automatic resource configuration.
Works about cloud reasoning are focused on other aspects like security attacks~\cite{choi2019ontology}, minimising sources like computing nodes~\cite{ghetas2018resource}, computational requirements~\cite{rakib2019efficient}, service placement~\cite{forti2022declarative}, verifying policy properties~\cite{backes2018semantic}, deploying semantic reasoning on the cloud~\cite{su2018distribution}. There is insufficient discussion on helping users to automate the cloud resource configuration.

\smallskip
\noindent 
\textbf{Uptake by stakeholders and benefits.}
Semantic technologies play an increasingly important role in modern industrial practice. Ontologies, as a good way for formal description of knowledge, offer unambiguous ``lingua franca'' for cross-domain communication. They can help users to perform tasks of a remote domain that otherwise would be error-prone, time-consuming and cognitively demanding. We incorporated rule-based reasoning, falling in the category of symbolic reasoning, with machine learning, which is a type of sub-symbolic reasoning. The combination takes benefits from both: \SemCloud becomes more agnostic of cloud infrastructure and adapts to the resource conditions, thus exploiting explicit domain knowledge via semantics and learning implicit relationship via ML. 

In addition, we tested \SemCloud with users of various backgrounds (welding experts, data scientists, semantic experts). \SemCloud could improve their working efficiency. 
Before using \SemCloud, users that are the non-cloud experts have very limited understanding of the cloud system, and did not use the cloud system.
Through \SemCloud, these users could obtain better understanding of the cloud system, start using the cloud system, and rely on the \SemCloud to automatically configure the resource allocation. We tested the GUI with the users to collected feedback for improving  the usability   and expanding the functionalities.
\looseness=-1

\smallskip
\noindent 
\textbf{Lessons Learnt on costs and risks.}
The main costs for development of such systems comprise the early development time for the semantic infrastructure that mediates between the cloud resources, data analysis solutions and users. Naturally, these costs vary depending on the specific project. It was manageable in our case, but should be carefully evaluated for each project individually. The key lessons learnt for reducing costs is that a good cross-domain communication framework is essential, where experts of different backgrounds can speak  a common language and reduce misunderstanding and communication time.
A possible and important risk is that the assumption could be wrong as to whether and to what extent the ETL and data analysis can be parallelised.
% Organisations need to evaluate which computational tasks can be assumed independent and which cannot.
% This evaluation needs to be performed early to avoid further costs.
It is recommended to verify the assumption early to avoid further costs.

\section{Conclusion,  and Outlook}
\label{sec:conclusion}

\noindent
\textbf{Conclusion.}
This work presents our \SemCloud system motivated by
a Bosch use case of welding monitoring, for addressing the scalability challenges in terms of \textit{data volume}, \textit{variety}, and more \textit{users}.
\SemCloud provides semantic abstraction that mediates between the users, ETL and data analysis, as well as cloud infrastructure. 
The scalability in terms of data variety is addressed by semantic data integration,  data volume by distributed ETL and data analysis, and scalability to more users by adaptive Datalog rule-based resource configuration.
% an ontology-based GUI system and adaptive Datalog rules.
These Datalog rules are adaptive because they have parameterised functions learnt from pilot running statistics via ML, a combination of symbolic and sub-symbolic approaches.
% The rules are then used for generating the cloud resource configuration automatically.
We evaluated \SemCloud extensively through cloud deployment with large volume of industrial data, and rule learning from thousands of pilot runs of the system, showing very promising results.
\looseness=-1
% rule learning and inference, and deployment of scaled ML solutions, which demonstrated very promising results.

\medskip
\noindent \textbf{Outlook.}
\SemCloud is under the umbrella of Neuro-Symbolic AI for Industry 4.0 at Bosch~\cite{DBLP:conf/semweb/ZhouTZZHZYT0GK22} that aims at enhancing manufacturing with both symbolic AI (such as semantic technologies~\cite{yahya2023semantic}) for improving transparency~\cite{zheng2022executable}, and ML for prediction power~\cite{2023tankgewelding}.
Bosch is developing user-friendly cloud technology in the framework of EU project DataCloud with many EU partners~\cite{datacloud}. \SemCloud is partially developed with production end-users and current deployed in a Bosch evaluation environment. We plan to  push it into the production to test with more users and collect feedback, and work together with EU partners for transferring the knowledge and experience to other manufacturing domains to increase the wide adoption.
We also plan to develop formal theories for knowledge representation and reasoning for cloud technology and automatic resource configuration, e.g., better  modelling framework, advanced reasoning rules, and deeper integration of symbolic and sub-symbolic reasoning.

% \paragraph*{Supplemental Material Availability:} 
% We open-source the SemCloud ontology under the link: \url{https://github.com/nsai-uio/SemCloud}~\cite{semcloudontology}. Other materials, such as the welding datasets, ML scripts are subject to corporate restrictions and are thus not provided.
% In addition, we plan to unfold the potential of mapping to transform data (meaningful portions of it) into linked data, i.e. to make \textit{Ontology Interpreter}  semantic by adding competency questions and visualisation graphs related to deep exploration of data, underlying ML models.

% Ontology can develop into a standard way of knowledge organisation in engineering. As can be seen from the feedback from domain experts, having learned the template technology, they can easily generalise the approach to other manufacturing processes. This makes different processes from the view of physics comparable from the view of ontology and data science.

% Ontology can develop into a standard way for bridging engineering and data science. The advantages of unambiguity, standardisation, and machine readability are highly desired features for data science.

\medskip
\noindent \textbf{Acknowledgements.}
The work was partially supported by the European Commission funded projects 
DataCloud (101016835),
enRichMyData (101070284),
Graph-Massivizer (101093202),
Dome 4.0 (953163), 
OntoCommons (958371), 
and the Norwegian Research Council funded projects (237898, 323325, 309691, 309834, and 308817).

\bibliographystyle{elsarticle-num}
\bibliography{iswc2023}
\end{document}